\documentclass{article}

\usepackage{PRIMEarxiv}

\usepackage[utf8]{inputenc} 
\usepackage[T1]{fontenc}    
\usepackage{hyperref}       
\usepackage{url}            
\usepackage{booktabs}       
\usepackage{amsfonts}       
\usepackage{nicefrac}       
\usepackage{microtype}      
\usepackage{lipsum}
\usepackage{fancyhdr}       
\usepackage{graphicx}       

\usepackage[numbers]{natbib}
\usepackage{amsmath}
\usepackage{xcolor}
\usepackage{amsthm}

\usepackage{algorithm}
\usepackage{algorithmic}
\usepackage{bbm}
\usepackage{adjustbox}
\usepackage{subcaption} 
\usepackage{multirow}
\usepackage{bm}

\allowdisplaybreaks

\title{Monitor and Recover: A Paradigm for Future Research on Distribution Shift in Learning-Enabled Cyber-Physical Systems}

\author{
    Vivian Lin, Insup Lee \\
    University of Pennsylvania \\
    Philadelphia, Pennsylvania, USA \\
    \texttt{\{vilin, lee\}@seas.upenn.edu}
}

\begin{document}

\maketitle

\begin{abstract}
With the known vulnerability of neural networks to distribution shift, maintaining reliability in learning-enabled cyber-physical systems poses a salient challenge. In response, many existing methods adopt a \textit{detect and abstain} methodology, aiming to detect distribution shift at inference time so that the learning-enabled component can abstain from decision-making. This approach, however, has limited use in real-world applications. We instead propose a \textit{monitor and recover} paradigm as a promising direction for future research. This philosophy emphasizes 1) robust safety monitoring instead of distribution shift detection and 2) distribution shift recovery instead of abstention. We discuss two examples from our recent work.
\end{abstract}

\section{Introduction}
Cyber-physical systems with learning-enabled components have become increasingly common, as data-driven approaches, notably neural networks, enable practitioners to tackle highly complex systems. However, reliable neural network performance is impeded by distribution shift~\citep{hendrycks2016baseline,fu2019diagnosing}, or differences between the distributions of data at training and inference time. This challenge that plagues standalone neural networks also affects learning-enabled cyber-physical systems (LE-CPS). When integrated with a cyber-physical system, errors can compound and have drastic consequences on the physical world with which the system interacts.

A variety of approaches have been proposed to mitigate the effects of distribution shift. Many attempt to diversify the training distribution through techniques like data augmentation~\citep{hendrycks2019augmix,yun2019cutmix} and domain generalization~\citep{gulrajani2020search,zhou2021domain}. Domain adaptation instead aims to train a model that generalizes to a target domain~\citep{bousmalis2017unsupervised,hoffman2018cycada}. While these approaches may improve a network's ability to generalize, the system could always encounter distributions not previously accounted for. Thus, inference time methods that dynamically respond to inputs are desirable. Common such methods are \textit{detect and abstain} approaches, which detect distribution shifts so that the network can abstain from making decisions on affected data~\citep{hendrycks2016baseline,kaur2022idecode,rabanser2019failing,zisselman2020deep}. However, this can be conservative, and needlessly flagging distributions against which the system is robust can have adverse effects. For example, similar hyper-conservatism in medical monitors has been shown to induce alarm fatigue, endangering patients~\citep{pugh2022evaluating}. The \textit{detect and abstain} approach also leads to inaction, an outcome that is highly undesirable when there is no human in the loop to assist. These practical constraints necessitate a new approach.

\section{Monitor and Recover}
We propose a \textit{monitor and recover} paradigm that entails two important directions for future work in LE-CPS research. \textit{Monitor}: robust safety monitoring may overcome the challenges of distribution shift detection. Since distribution shift does not always lead to failure, directly monitoring the safety properties of LE-CPS is less conservative.
Of course, the method of safety monitoring must itself maintain strong performance under distribution shift. \textit{Recover}: dynamic modification of the incoming data may enable reliable decision-making by the learning-enabled component. By avoiding abstention, this distribution shift recovery allows LE-CPS to act. We offer two examples from our recent work, positioning them among related literature following the \textit{monitor and recover} paradigm.

\begin{figure*}[!t]
    \begin{subfigure}[h]{0.5\linewidth}
        \includegraphics[width=\linewidth]{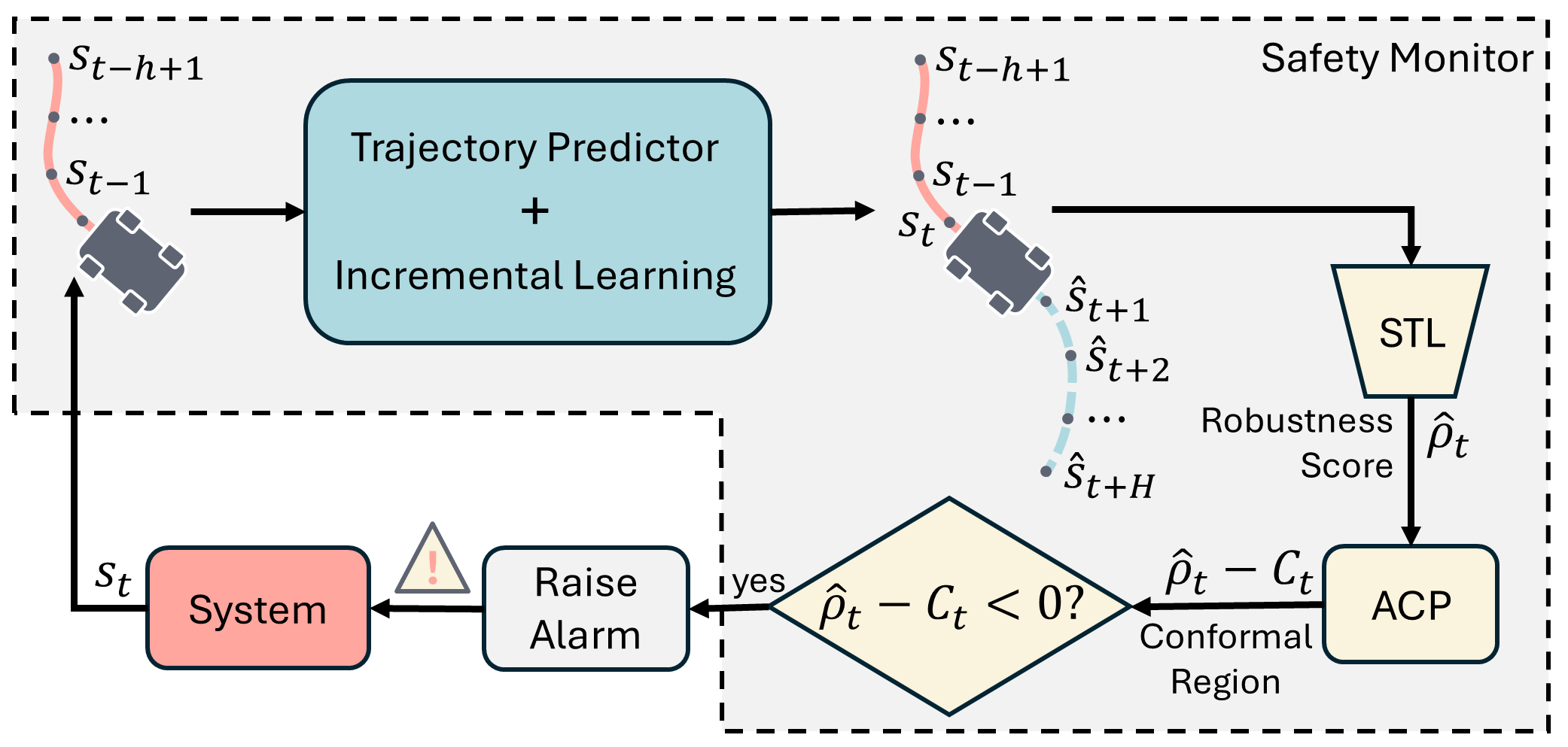}
        \caption{Robust Safety Monitoring\label{fig:monitor}}
    \end{subfigure}
    \hspace{0.03\linewidth}
    \begin{subfigure}[h]{0.4\linewidth}
        \includegraphics[width=\linewidth]{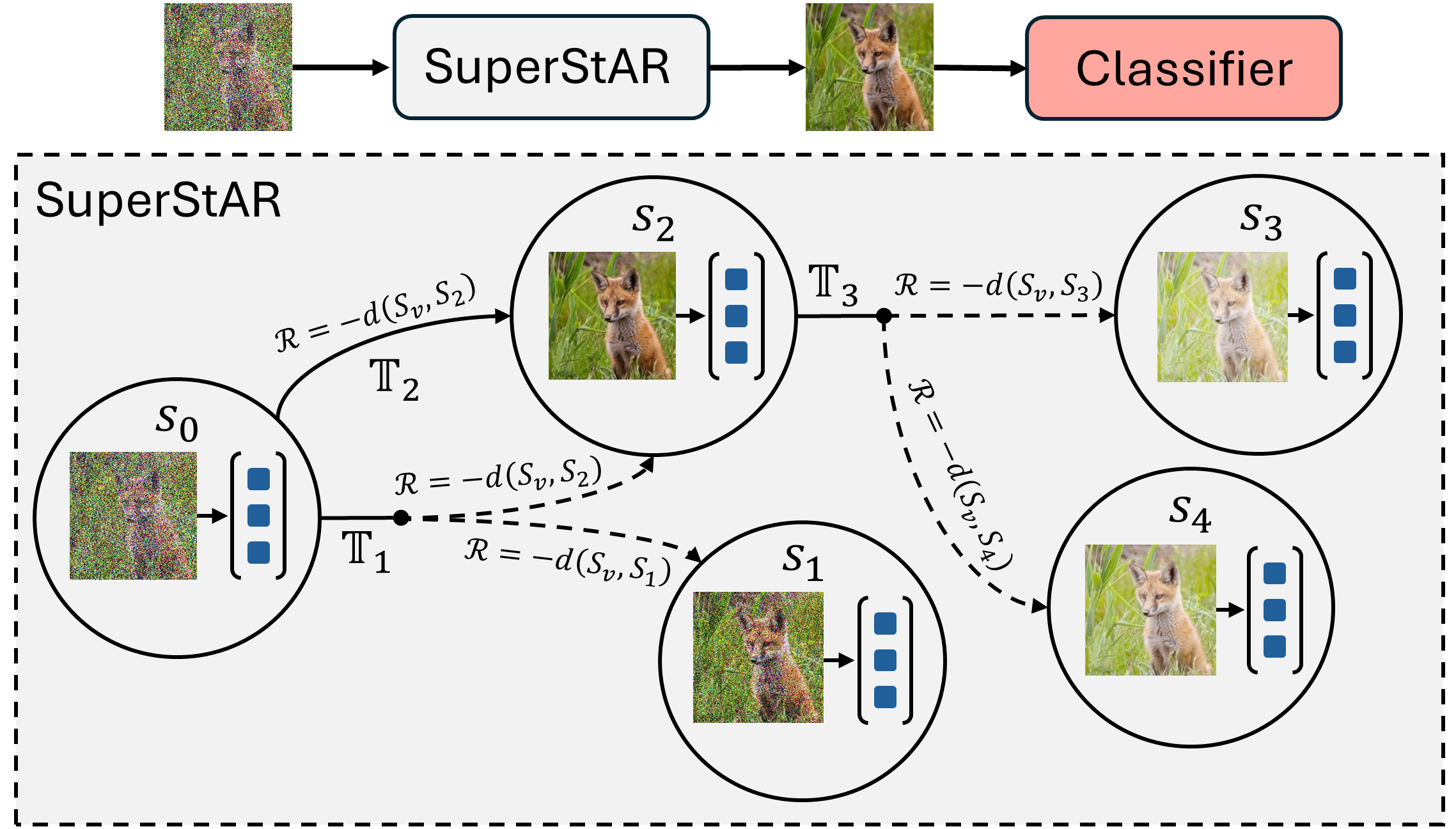}
        \caption{Distribution Shift Recovery\label{fig:recover}}
    \end{subfigure}%
    \caption{Example recent works following the \textit{monitor and recover} paradigm. a) An example of robust safety monitoring~\citep{lin2025safety}. A trajectory predictor equipped with incremental learning predicts trajectories of the system states. On this prediction, a conformal region over the STL robustness score is computed using adaptive conformal prediction (ACP). A simple check indicates whether a safety violation is predicted to occur. b) Distribution shift recovery via SuperStAR~\citep{lin2024dc4l}. A sequence of transforms bring the data closer to the training distribution. The transforms are selected by a reinforcement learning agent. The data is represented by state $S_i$ and undergoes transforms $\mathbb{T}_j$ with reward $\mathcal{R}$ defined by the distance $d$ between the validation state $S_v$ and the current state $S_i$.}
\end{figure*}

The runtime safety monitor proposed in~\citet{lin2025safety} predicts violations of signal temporal logic (STL) safety properties. This is achieved by calculating the STL robustness value (i.e., the distance from property satisfaction) of predicted system state trajectories. A prediction region is generated over this robustness value using adaptive conformal prediction, which enables probabilistic coverage guarantees under \textit{any} distribution shift. Competing methods assume no~\citep{lindemann2023conformal} or bounded~\citep{zhao2024robust} shift. Incremental learning is also incorporated to limit the conservatism that naturally results from interval predictions. Figure~\ref{fig:monitor} shows an overview of this approach. Empirical evaluations were performed on simulated autonomous vehicles in the presence of static and dynamic obstacles. The safety monitor raises violation alarms in a timely manner with high recall even when distribution shift occurs, and the addition of incremental learning increases precision. With further improvements to computational efficiency, this method can be deployed 
in real-time.

The distribution shift recovery approach from \citet{lin2024dc4l} returns inference-time data to the training distribution through a sequence of semantic-preserving transforms. The dynamic selection of these transforms is formulated as a Markov decision process, where the actions (i.e., transforms) are chosen based on a state representation of the data and the Wasserstein distance between a test and validation set. The task can then be solved by reinforcement learning. An overview of the technique is shown in Figure~\ref{fig:recover}. Unlike test time augmentation approaches~\citep{guo2017countering,lyzhov2020greedy}, this method dynamically selects transforms without querying the downstream model. Retraining or fine-tuning the downstream model, typical in test time adaptation~\citep{gandelsman2022test,wang2022continual}, are also not required. Evaluations were performed for image classifiers under naturally occurring distribution shift, such as weather changes and image corruptions. By applying appropriate sequences of image transformations, including denoising methods and histogram equalization, the technique partially recovers classifier performance under distribution shift.

\section{Conclusions}
\textit{Detect and abstain} approaches for addressing distribution shift in learning-enabled cyber-physical systems face practical challenges. These methods lead to conservatism and inaction that hinder deployment in the real world. We posit instead that a paradigm of \textit{monitor and recover} is a promising direction of future research in this area. Such techniques emphasize direct monitoring of safety properties and recovery of actionable data under distribution shift.


\section*{Acknowledgements}
This work was supported in part by ARO MURI W911NF-20-1-0080, NSF 2143274, and a gift from AWS AI to Penn Engineering's ASSET Center for Trustworthy AI. Any opinions, findings, conclusions or recommendations expressed in this material are those of the authors and do not necessarily reflect the views the Army Research Office (ARO), the Department of Defense, or the United States Government.

\bibliographystyle{plainnat}
\bibliography{ref}

\end{document}